\documentclass[a4paper, 10pt, conference]{IEEEtran}
\IEEEoverridecommandlockouts
\usepackage{hyperref}
\usepackage[T1]{fontenc}
\usepackage{amsmath,amssymb,amsfonts,empheq}
\usepackage{graphicx}
\usepackage{float}
\usepackage{textcomp}
\usepackage{xcolor}
\usepackage{subcaption}
\usepackage{algorithm,algcompatible}
\usepackage{amssymb}
\algnewcommand\algorithmicreturn{\textbf{return}}
\usepackage[backend=biber,style=ieee]{biblatex}
\usepackage{biblatex}

\newcommand{\mb}{\mathbf}

\begin{document}

\title{\LARGE \bf Whole-body MPC and sensitivity analysis of a real time foot step sequencer for a biped robot Bolt}

\author{Constant Roux$^1$\thanks{$^1$Gepetto Team, LAAS-CNRS, Universit\'e de Toulouse, France.}, Côme Perrot$^1$, Olivier Stasse$^{1,2}$\thanks{$^2$Artificial and Natural Intelligence Toulouse Institute, France, {firstname.surname@laas.fr}}}


\maketitle

\begin{abstract}
This paper presents a novel controller for the bipedal robot Bolt.
Our approach leverages a whole-body model predictive controller in conjunction with a footstep sequencer to achieve robust locomotion.
Simulation results demonstrate effective velocity tracking as well as push and slippage recovery abilities.
In addition to that, we provide a theoretical sensitivity analysis of the footstep sequencing problem to enhance the understanding of the results. 
\end{abstract}


\section{Introduction}
\subsection{Context}
Bipedal robotics, with its origins tracing back to the end of the last century, has witnessed a significant surge in recent years.
This trend, driven by technological advancements in areas such as actuation, and computing, has opened up a flurry of new potential applications \cite{unitree_h1_2024, fourier_gr1_2024}.
However, these new solutions require efficient controllers that can fully exploit the hardware's specificities to maximize utility.


Historically, trajectory optimization (TO) has been widely used for the control of bipedal robots. 
Early models, such as the Linear Inverted Pendulum (LIP) or centroidal Model Predictive Control (centroidal MPC), evolved into more sophisticated approaches like whole-body MPC (WB MPC) \cite{Wensing_tro_2024, assirelli:hal-03778738, li_cafe-mpc_2024, dantec_whole_2021}.
However, as the complexity of control problems increases, the limitations of MPC become apparent. 
These limitations include limited reactivity to unforeseen disturbances and insufficient computational capacity.
Even though warm starting the solver near the optimal solution can drastically reduce computation time \cite{lembono2020memory}, these challenges persist. 

Reinforcement Learning (RL) techniques have emerged as a promising alternative \cite{singh_access_2023, zhuang_humanoid_2024, radosavovic_real_world_2024}.
RL nonetheless presents notable limitations. 
Bipedal robots are inherently very unstable, complicating the learning process. 
Additionally, RL requires vast amounts of data and computational time, and often lacks performance guarantees in real-world conditions. 
RL algorithms can also be sensitive to variations in the training environment, limiting their robustness and generalizability.

In response to these challenges, hybrid methods, combining the strengths of trajectory optimization and reinforcement learning, are gaining popularity. 
Approaches such as Deep Tracking Control (DTC) \cite{jenelten:sr:2024} or Actor-Critic
Model Predictive Control \cite{romero_actor_critic_2024} aim to merge the best of both worlds to achieve optimal performance in terms of stability and adaptability.
The enthusiasm around hybrid methods highlights the relevance of traditional MPC methods, especially when dealing with highly unstable humanoids.
However, in most applications, they require an additional trajectory planner, which may not account for the dynamics of the MPC, to provide a reference trajectory for the robot.

\begin{figure}[t]
    \vspace*{-\intextsep}
    \centering
    \includegraphics[width=\columnwidth]{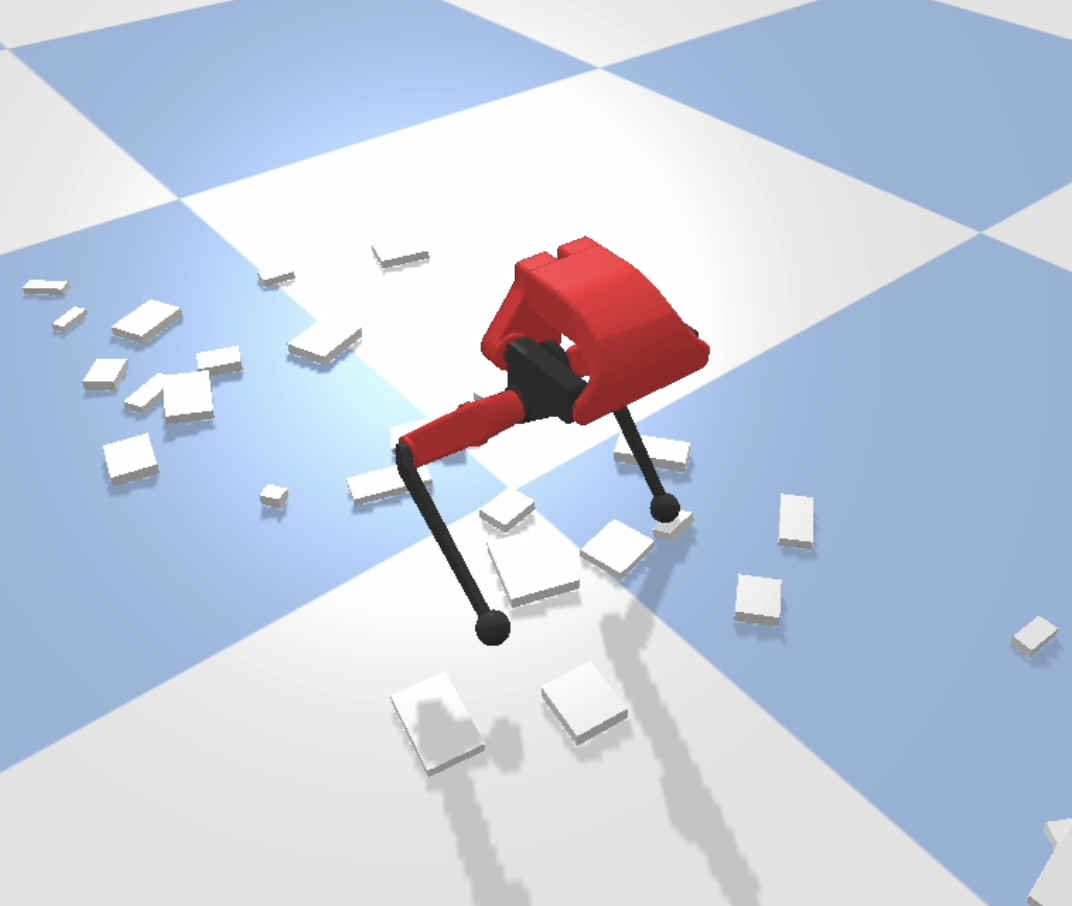}
    \caption{Bipedal robot bolt walking in PyBullet using whole-body MPC on cluttered terrain.}
    \label{fig:bolt_cluttered_terrain}
\end{figure}

Despite increasingly precise modeling and step sequencers based on simplified yet effective models, such as the Divergent Component of Motion (DCM) \cite{englsberger_three-dimensional_2015, zhang_robust_2022}, these planners are still necessary to provide robust heuristics for bipedal walking.
An example of such architecture is presented in \cite{khadiv_itro_2020}.
It employs a step planner followed by a foot trajectory controller and a whole-body instantaneous controller.
Similarly, \cite{Yeganegi_2022} conducted research using a low-frequency step planner (10 Hz) and a whole-body controller.
A method based on iterative Linear Quadratic Gaussian (iLQG) \cite{Tassa_2012} provides the capabilities to find foot sequences autonomously, but struggles to efficiently work on real hardware \cite{Koenemann_2015}.
However, DDP with a rigid contact formulation such as the one formulated in \cite{mastalli20crocoddyl} and tested on TALOS in \cite{dantec_whole_2021} shows that if a sequence of contacts is given, WB MPC is doable on a real size humanoid robot.

\subsection{Contributions}

In this paper, we propose to build upon the work of Khadiv et al. \cite{khadiv_itro_2020} by replacing the combination of the whole-body instantaneous controller and the footstep trajectory controller with a unified whole-body trajectory controller.
Our approach aims to operate both the step sequencer and the whole-body model predictive controller (WB MPC) at the same frequency of 100 Hz, contrary to Yeganegi et al. \cite{Yeganegi_2022}.
This approach leads to the emergence of footstep trajectories, as demonstrated in previous works \cite{Dantec:ral:2022, assirelli:hal-03778738}, thereby simplifying robot programming and explicitly accounting for the angular momentum induced by foot motions.

Furthermore, we propose a parameter sensitivity analysis of the step sequencer based on the method proposed by Fiacco \cite{fiacco_1983}.
The primary aim of this analysis is to provide analytical validation of intuitions regarding the interpretation of results.
Secondly, it will serve as a preliminary basis for future research.

The contributions of this work are as follows:
\begin{itemize}
    \item Integration of a step sequencer with a fixed-horizon WB MPC framework.
    \item Simulation of the Bolt robot in PyBullet 
    under various perturbation scenarios to evaluate the control strategy.
    \item Comprehensive sensitivity analysis of the step sequencer's response to disturbances in the DCM.
\end{itemize}

\section{Footsteps Sequencer} \label{sec:DCM-MPC}
\label{section:footsteps_sequencer}
\subsection{Review of previous work}
Khadiv et al. \cite{khadiv_itro_2020} proposed to compute the position and timing of the next step so that the robot's center of mass adheres to a desired velocity command while stabilizing the DCM.
The DCM is the unstable component of the linear inverted pendulum model, whose dynamic equations are recalled as follows:
\begin{subequations}
    \label{equ:COM-DCM-system}
    \begin{align}
        \dot{\mb{c}} &= w_0(\mb{\zeta}-\mb{c}) \label{equ:CoM-dynamics}  \\
        \dot{\mb{\zeta}} &= w_0(\mb{\zeta}-\mb{p}_0) \label{equ:DCM-dynamics}
    \end{align}
   
\end{subequations}
where \( \mb{c} \in \mathbb{R}^2 \) is the position of the robot's center of mass, \( \omega_0 \in \mathbb{R} \) is the natural frequency of the pendulum (\( \omega_0 = \sqrt{\frac{g}{z_c}} \), with \( g \) being the gravitational constant and \( z_c \) the fixed height of the center of mass), \( \mb{\zeta} \in \mathbb{R}^2 \) is the robot's DCM, and \( \mb{p}_0 \in \mathbb{R}^2 \) is the position of the support foot.

To achieve this, Khadiv et al. \cite{khadiv_itro_2020} propose solving the following linearly constrained multi-objective QP problem:
\begin{equation}
    \label{equ:QP-problem}
    \begin{aligned} 
        \min _{\mathbf{x}} & \quad f(\mathbf{x}) & \\ 
    \text { s.t. } & 
    \begin{cases}
        g_i(\mathbf{x}) \geq 0, & \quad i = 1, \ldots, 6  \\ 
        \hat{h_j}(\mathbf{x}) = 0, & \quad j = 1, \ldots, 2
    \end{cases}
    \end{aligned}
\end{equation}
where \(\mathbf{x}=\begin{pmatrix} \mb{p}_T^{\top} & \Gamma(T) & \mb{b}_T^{\top} \end{pmatrix}^{\top}\) with \(\mb{p}_T=\begin{pmatrix}p_{T,x} & p_{T,y}\end{pmatrix}^{\top}\ \in \mathbb{R}^2\) representing the position of the next step, \( \Gamma(T)=e^{w_0 T} \in \mathbb{R} \) the timing of the next step's contact, and \( \mathbf{b}_T \in \mathbb{R}^2\) the DCM offset at the next step's contact instant (\( \mathbf{b}_T=\begin{pmatrix}b_{T,x} & b_{T,y}\end{pmatrix}^{\top}= \mb{\zeta}_T - \mathbf{p}_T \), where \( \mb{\zeta}_T \) is the DCM at the next step's contact instant). The multi-objective function is given as follows:
\begin{subequations}
    \label{equ:f-func}
    \begin{align}
    f(\mathbf{x}) = &\ \alpha_1 \left \| \mathbf{p}_T - \mathbf{p}_0 - \begin{pmatrix} 
            l_{\text{nom}}\\ 
            w_{\text{nom}}
            \end{pmatrix} \right \|^2 \notag \\
        & + \alpha_2 \left | \Gamma(T) - \Gamma(T_{\text{nom}}) \right |^2 \notag \\
        & + \alpha_3 \left \| \mathbf{b}_T - \begin{pmatrix}
            b_{x,\text{nom}}\\ 
            b_{y,\text{nom}}
            \end{pmatrix} \right \|^2
    \end{align}
\end{subequations}
with \( \alpha_1 \), \( \alpha_2 \), \( \alpha_3 \) as the weights of the different objectives, \( l_{\text{nom}} \) and \( w_{\text{nom}} \) as the desired step length and width respectively, \( \Gamma(T_{\text{nom}}) \) the desired contact time, and \( b_{x,\text{nom}} \), \( b_{y,\text{nom}} \) the desired DCM offsets whose analytical expression is available in Appendix A of \cite{khadiv_itro_2020}. The inequality constraints are given as follows:
\begin{subequations}
    \label{equ:inequality-constraints}
    \begin{equation}
    \label{equ:inequality-constraints-1}
    g_1(\mathbf{x}) = p_{T,x} - p_{0,x} - l_{\text{min}}
    \end{equation}
    \begin{equation}
    \label{equ:inequality-constraints-2}
    g_2(\mathbf{x}) = l_{\text{max}} - p_{T,x} + p_{0,x}
    \end{equation}
    which are constraints limiting the step length between \( l_{\min} \) and \( l_{\max} \),
    \begin{equation}
    \label{equ:inequality-constraints-3}
    g_3(\mathbf{x}) = p_{T,y} - p_{0,y} - w_{\text{min}}
    \end{equation}
    \begin{equation}
    \label{equ:inequality-constraints-4}
    g_4(\mathbf{x}) = w_{\text{max}} - p_{T,y} + p_{0,y}
    \end{equation}
    which are constraints limiting the step width between \( w_{\min} \) and \( w_{\max} \),
    \begin{equation}
    \label{equ:inequality-constraints-5}
    g_5(\mathbf{x}) = \Gamma(T) - \Gamma({T_{\text{min}}})
    \end{equation}
    \begin{equation}
    \label{equ:inequality-constraints-6}
    g_6(\mathbf{x}) = \Gamma({T_{\text{max}}}) - \Gamma(T)
    \end{equation}
    which are constraints limiting the step contact time between \( \Gamma(T_{\min}) \) and \( \Gamma(T_{\max}) \).
\end{subequations}

The following equations are equality constraints ensuring that the DCM follows the dynamics imposed by Eq. \eqref{equ:DCM-dynamics}:
\begin{subequations}
    \label{equ:equality-constraints}
    \begin{align}
    \label{equ:equality-constraints-1}
    \hat{h_1}(\mathbf{x}) &= p_{T,x} + b_{T,x} - p_{0,x} - (\hat{\zeta}_x - p_{0,x}) e^{-w_0 t} \Gamma(T) \\ 
    \label{equ:equality-constraints-2}
    \hat{h_2}(\mathbf{x}) &= p_{T,y} + b_{T,y} - p_{0,y} - (\hat{\zeta}_y - p_{0,y}) e^{-w_0 t} \Gamma(T)
    \end{align}
\end{subequations}
with \( t \) the time elapsed since the last foot contact and \( \hat{\mb{\zeta}}\) the measured DCM.
Note that, as explained in \cite{Daneshmand_2021}, the DCM offset is not hard constrained because this could make the problem infeasible under certain conditions. In addition, \(w\) will sometimes be referred to as \(w_{\text{left}}\) or \(w_{\text{right}}\) depending on whether the next step is executed by the left foot or the right foot.

\subsection{Sensitivity analysis to disturbances on the measured DCM} \label{subsec:sensitivity_analysis}
We now propose to analyze the sensitivity of the optimal solution to perturbations on the measured DCM. These perturbations can model measurement noise or an external force that would alter the value of the DCM.

Introducing a disturbance on the measured DCM such that \( \hat{\mb{\zeta}} = \mb{\zeta} + \mb{\theta} \), where \( \mb{\zeta} \in \mathbb{R}^2 \) is the real value of the DCM and \( \mb{\theta} \in \mathbb{R}^2 \) is a disturbance, the problem \eqref{equ:QP-problem} then becomes:
\begin{equation}
    \label{equ:QP-problem-disturbance}
    \begin{aligned} 
        \min _{\mathbf{x}} & \quad f(\mathbf{x}) & \\ 
    \text { s.t. } & 
    \begin{cases}
        g_i(\mathbf{x}) \geq 0, & \quad i = 1, \ldots, 6  \\ 
        h_j(\mathbf{x}) + \theta_j c_j(\mathbf{x}) = 0, & \quad j = 1, \ldots, 2
    \end{cases}
    \end{aligned}
\end{equation}

\begin{subequations}
    \label{equ:equality-constraints-perturbation}
    \begin{align}
    \label{equ:equality-constraints-perturbation-1}
    h_1(\mathbf{x}) &= p_{T,x} + b_{T,x} - p_{0,x} - (\zeta_x - p_{0,x}) e^{-w_0 t} \Gamma(T) \\ 
    \label{equ:equality-constraints-perturbation-2}
    h_2(\mathbf{x}) &= p_{T,y} + b_{T,y} - p_{0,y} - (\zeta_y - p_{0,y}) e^{-w_0 t} \Gamma(T)
    \end{align}
\end{subequations}

\begin{subequations}
    \label{equ:equality-constraints-c}
    \begin{align}
    \label{equ:equality-constraints-c1}
    c_1(\mathbf{x}) &= -e^{-w_0 t} \Gamma(T) \\ 
    \label{equ:equality-constraints-c2}
    c_2(\mathbf{x}) &= -e^{-w_0 t} \Gamma(T)
    \end{align}
\end{subequations}

The theoretical result of sensitivity of a QP problem with respect to a parameter is provided by \cite{fiacco_1983}. Subsequently, the use of the superscript \(^*\) on an expression will denote that the latter is evaluated at \(\mb{\theta}=0\).
Noting that (a) the functions of Equation \eqref{equ:QP-problem-disturbance} are twice differentiable, (b) the second-order sufficiency conditions hold at \( \mathbf{x}^* \), (c) \( \{ \nabla_{\mathbf{x}}g^*_i \}, i=1, \dots, 6 \), \( \{ \nabla_{\mathbf{h}}g^*_j \}, j=1, \dots, 2 \) are linearly independent, and (d) \( u_i^* > 0\) when \( g_i^*(\mathbf{x})=0\) where \( u_i\) and \( w_i \) are the Lagrangian multipliers, then the following set of equations is satisfied at \((\mathbf{x}, \mathbf{u}, \mathbf{w})=(\mathbf{x}^*, \mathbf{u}^*, \mathbf{w}^*)\), \( \mb{\theta}=0 \):
\begin{equation}
    \label{equ:KKT}
    \begin{aligned}
    & \begin{aligned}[t]
    & \nabla_{\mathbf{x}}f - \sum_{i=1}^{6} u_i \nabla_{\mathbf{x}}g_i + \sum_{j=1}^{2} w_j (\nabla_{\mathbf{x}}h_j + \theta_j \nabla_{\mathbf{x}}c_j) = 0
    \end{aligned} \\
    & \begin{aligned}[t]
    & u_i g_i(\mathbf{x}) = 0, && i = 1, \ldots, 6 \\
    & h_j(\mathbf{x}) + \theta_j c_j(\mathbf{x}) = 0, && j = 1, 2
    \end{aligned}
    \end{aligned}
\end{equation}

Let \( F(\mathbf{x}, \mb{\theta}) \) denote the function composed of the terms on the left-hand side of Eq. \ref{equ:KKT}. Applying the implicit function theorem, we deduce that:
\begin{equation}
    \label{equ:analytical-QP-sensitivity}
    \frac{\partial (\mathbf{x}^*,\mathbf{u}^*,\mathbf{w}^*)}{\partial \theta}=-J_{F^*}^{-1}(\mathbf{x^*},\mathbf{u^*},\mathbf{w^*})J_{F}(\theta)
\end{equation}

\begin{equation}
    J_{F^*}(\mathbf{x^*})=
    \begin{pmatrix}
    \text{diag}(\alpha_i) & -G^* & H^*\\ 
    
    \text{diag}(u_i^*)(G^*)^\top & \text{diag}(g_i^*) & [0]_{6 \times 2} \\ 
    
    (H^*)^\top & [0]_{2 \times 6} & [0]_{2 \times 2}
    \end{pmatrix}
\end{equation}

\begin{equation}
    J_{F}(\theta)=
    \begin{pmatrix}
    C^*\text{diag}(w_j^*)\\ 
    [0]_{6 \times 2}\\ 
    \text{diag}(c_j^*)
    \end{pmatrix}
\end{equation}

where \( G^* = \begin{pmatrix} \nabla_\mathbf{x}g_1^* & \cdots & \nabla_\mathbf{x}g_6^* \end{pmatrix} \), \( H^* = \begin{pmatrix} \nabla_\mathbf{x}h_1^* & \nabla_\mathbf{x}h_2^* \end{pmatrix} \) and \( C^* = \begin{pmatrix} \nabla_\mathbf{x}c_1^* & \nabla_\mathbf{x}c_2^* \end{pmatrix} \).

\begin{figure}[t]
\centering
  \begin{subfigure}{.48\columnwidth}
    \includegraphics[width=\linewidth]{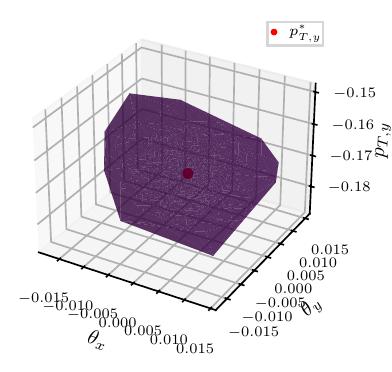}
  \end{subfigure}
  \begin{subfigure}{.48\columnwidth}
    \includegraphics[width=\linewidth]{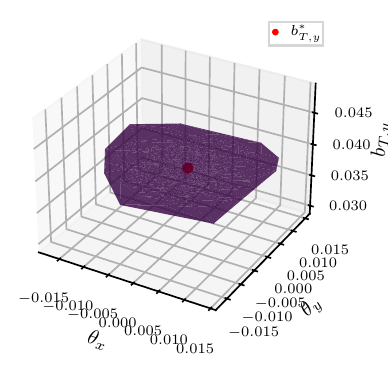}
  \end{subfigure}
\caption{Solution space example of \( p_{T,y} \) (left) and \( b_{T,y} \) (right) as a function of the DCM disturbance \( \theta \sim \mathcal{N}(0, 0.005) \) generated with 1000 samples.
The optimal solution is represented in red, and no inequality constraints is active.}
\label{fig:distrubance-space-comparison}
\end{figure}

Eq. \eqref{equ:analytical-QP-sensitivity} allows us to numerically evaluate the sensitivity of the optimal solution of the QP problem with respect to perturbations on the DCM.
For example, Fig.~\ref{fig:distrubance-space-comparison} presents the optimal solutions \( p_{T,y} \) and \( b_{T,y} \) in the space of Gaussian noise perturbations following \( \mathcal{N}(0, 0.005) \) where no inequality constraints are active, using the input parameters of the problem from Table \ref{tab:QP-params}.
We observe that these spaces form planes, and the slope of the plane in the space of \( p_{T,y} \) is steeper than that of \( b_{T,y} \). This result is expected because \( \alpha_3 \gg \alpha_1 \), which is also reflected in \( (\frac{{\partial p_{T,y}^*}}{{\partial\theta_y}} = 5.18) > (\frac{{\partial b_{T,y}^*}}{{\partial \theta_y}} = 5.18e-3) \), by using Eq. \eqref{equ:analytical-QP-sensitivity}.
Thus, we understand that the robot will prioritize balancing before responding to the speed command.

\begin{table}[t]
\centering
\caption{Parameters of the step sequencer used for walking}
\label{tab:QP-params}
\begin{tabular}{cccc}
\hline
Parameter                               & min       & nom               & max \\ \hline
($\alpha_1$, $\alpha_2$, $\alpha_3$)    & -         & (1e3, 1, 1e6)     & - \\
$z_c$                                   & -         & 0.31              & -     \\
$p_0$                                   & -         & (-0.12, 0.10)     & -     \\
$t$                                     & -         &   0.229           & -     \\
$\hat{\mb{\zeta}}$                           & -         & (-0.12, -0.07)    & -     \\
$l$                                     & -0.3      & 0.1               & 0.3  \\
$w_{\text{left}}$                       & -0.40     & -0.25             & -0.10  \\
$w_{\text{right}}$                      & 0.10      & 0.25              & 0.40 \\
$T$                                     & 0.1       & 0.3               & 1.0   \\ \hline
\end{tabular}
\end{table}

\subsection{Fixed-Horizon MPC Sequencer}
To enable the WB MPC to predict a state trajectory throughout its prediction horizon (see Section \ref{sec:WB-MPC}), the step sequence has to extend at least to the end of the MPC's horizon.
To generate a step sequence up to an appropriate time \( H_u \), we propose to iteratively solve problem~\eqref{equ:QP-problem} and reuse the solution computed at time \( k \) as the initial condition of the problem solved at time \( k+1 \).
This process can be carried out until a contact time exceeding the desired time horizon is obtained.
The problem can be summarized as follows:
\begin{equation}
    \label{equ:QP-problem-loopbacked}
    \begin{aligned} 
        \min _{\mathbf{x}_{k+1}} & \quad f(\mathbf{x}_{k+1}, \mathbf{x}_k) & \\ 
    \text { s.t. } & 
    \begin{cases}
        g_i(\mathbf{x}_{k+1}, \mathbf{x}_{k}) \geq 0, & \quad i = 1, \ldots, 6  \\ 
        h_j(\mathbf{x}_{k+1}, \mathbf{x}_{k}) = 0, & \quad j = 1, \ldots, 2
    \end{cases}
    \end{aligned}
\end{equation}

\begin{subequations}
    \label{equ:f-func-loopbacked}
    \begin{align}
    f(\mathbf{x}_{k+1}, \mathbf{x}_{k}) = &\ \alpha_1 \left \| \mb{p}_{T_{k+1}} - \mb{p}_{T_k} - \begin{pmatrix} 
            l_{\text{nom}}\\ 
            w_{k,\text{nom}}
            \end{pmatrix} \right \|^2 \notag \\
        & + \alpha_2 \left | \Gamma(T_{k+1}) - \Gamma(T_k + T_{\text{nom}}) \right |^2 \notag \\
        & + \alpha_3 \left \| \mb{b}_{T_{k+1}} - \begin{pmatrix}
            b_{k,x,\text{nom}}\\ 
            b_{k,y,\text{nom}}
            \end{pmatrix} \right \|^2
    \end{align}
\end{subequations}

\begin{subequations}
    \label{equ:inequality-constraints-loopbacked}
    \begin{align}
    \label{equ:inequality-constraints-1-lp}
    g_1(\mathbf{x}_{k+1}, \mathbf{x}_{k}) &= p_{T_{k+1},x} - p_{T_k,x} - l_{\text{min}} \\
    \label{equ:inequality-constraints-2-lp}
    g_2(\mathbf{x}_{k+1}, \mathbf{x}_{k}) &= l_{\text{max}} - p_{T_{k+1},x} + p_{T_k,x} \\
    \label{equ:inequality-constraints-3-lp}
    g_3(\mathbf{x}_{k+1}, \mathbf{x}_{k}) &= p_{T_{k+1},y} - p_{T_k,y} - w_{k,\text{min}} \\
    \label{equ:inequality-constraints-4-lp}
    g_4(\mathbf{x}_{k+1}, \mathbf{x}_{k}) &= w_{k,\text{max}} - p_{T_{k+1},y} + p_{T_k,y} \\
    \label{equ:inequality-constraints-5-lp}
    g_5(\mathbf{x}_{k+1}, \mathbf{x}_{k}) &= \Gamma({T_{k+1}}) - \Gamma({T_k + T_{\text{min}}}) \\
    \label{equ:inequality-constraints-6-lp}
    g_6(\mathbf{x}_{k+1}, \mathbf{x}_{k}) &= \Gamma({T_k + T_\text{max}}) - \Gamma({T_{k+1}})
    \end{align}
\end{subequations}

\begin{subequations}
    \label{equ:equality-constraints-loopbacked}
    \begin{align}
    \label{equ:equality-constraints-1-lb}
    h_1(\mathbf{x}_{k+1}, \mathbf{x}_k) &= p_{T_{k+1},x} + b_{T_{k+1},x} - p_{T_k,x} \nonumber \\ 
    & \quad - (\hat{\zeta}_{k,x} - p_{T_k,x}) e^{-w_0 T_k} \Gamma(T_{k+1}) \\ 
    \label{equ:equality-constraints-2-lb}
    h_2(\mathbf{x}_{k+1}, \mathbf{x}_k) &= p_{T_{k+1},y} + b_{T_{k+1},y} - p_{T_k,y} \nonumber \\
    & \quad - (\hat{\zeta}_{k,y} - p_{T_k,y}) e^{-w_0 T_k} \Gamma(T_{k+1})
    \end{align}
\end{subequations}
where \( \mathbf{x}_{k+1} \) represents the next solution and \( \mathbf{x}_k \) the previous solution. The parameter \( w_k \) varies depending on whether the next step is taken by the left foot or the right foot.

\begin{figure}[t]
\centering
    \includegraphics[width=\linewidth]{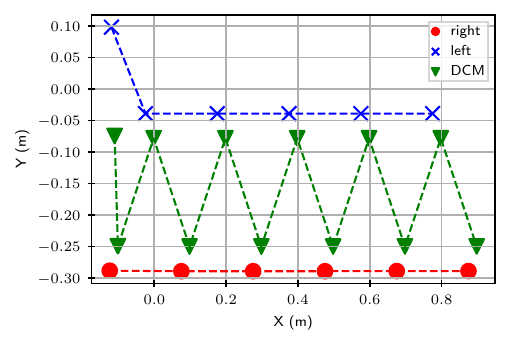}
\caption{Position of the feet and DCM in the Cartesian plane along a generated step sequence. 
The walking sequence was generated  with \(H_u=3\;s\) using the input parameters from Table \ref{tab:QP-params}. The imposed speed is \(V_x^*=\frac{l_{\text{nom}}}{T_{\text{nom}}}=0.33m.s^{-1}\). All constraints are satisfied, and the nominal speed is achieved.}
\label{fig:DCM-MPC-example}
\end{figure}

The following algorithm summarizes the MPC step sequencer:
\begin{algorithm}[H]
\caption{Footseps sequencer}
\begin{algorithmic}
\STATE $t \; \gets  t_{\text{mea}}$, $t_0 \; \gets \; t_{\text{mea}}$, $\hat{\mb{\zeta}}_k \; \gets \;  \mb{\zeta}_{\text{mea}}$
\STATE $\mathbf{x}_k \; \gets \; \begin{pmatrix} \mb{p}_{\text{ini}}^\top & \Gamma (t_{\text{mea}}) & [0]_{1 \times 2} \end{pmatrix}^\top$
\STATE $S \; \gets \; \emptyset$
\REPEAT
\STATE $\mathbf{x}_{k+1} \; \gets \; \text{Solve } \eqref{equ:QP-problem-loopbacked} \text{ using } \mathbf{x}_k$
\STATE $t \; \gets \; T_{k+1}$, $\hat{\mb{\zeta}}_k \; \gets \; \mb{p}_{T_{k+1}}+\mb{b}_{T_{k+1}}$
\STATE $S \; \gets \; S \cup \mathbf{x}_{k+1}$
\STATE $\mathbf{x}_k \; \gets \; \mathbf{x}_{k+1}$
\UNTIL{$t \geq t_0 + H_u$}
\STATE \textbf{Return } $S$
\end{algorithmic}
\end{algorithm}
where \( t_{\text{mea}} \) is the time at which the algorithm is called, \( \mb{\zeta}_{\text{mea}} \) is the measured DCM at \( t = t_{\text{mea}} \), and \( \mb{p}_{\text{ini}} \) is the position of the foot currently in support.
\(S\) is the list containing the \(K\) steps of the generated sequence.
The calculation of the nominal DCMs and the changes in \( w \) depending on whether the next step is taken by the left foot or the right foot are not detailed here for the sake of clarity.
Fig.~\ref{fig:DCM-MPC-example} illustrates an example of a step sequence generated using the parameters from Table \ref{tab:QP-params} and a horizon \(H_u\) of $3\;s$ for walking at a velocity of \(V_x^*=0.3\;m.s^{-1}\).
The figure also shows that all hard constraints are satisfied, and the DCM remains non-divergent over time.
This sequence can subsequently serve as a reference for WB MPC.

\section{Whole-body MPC} \label{sec:WB-MPC}

\begin{figure*}[t!]
    \centering
    \includegraphics[width=\textwidth]{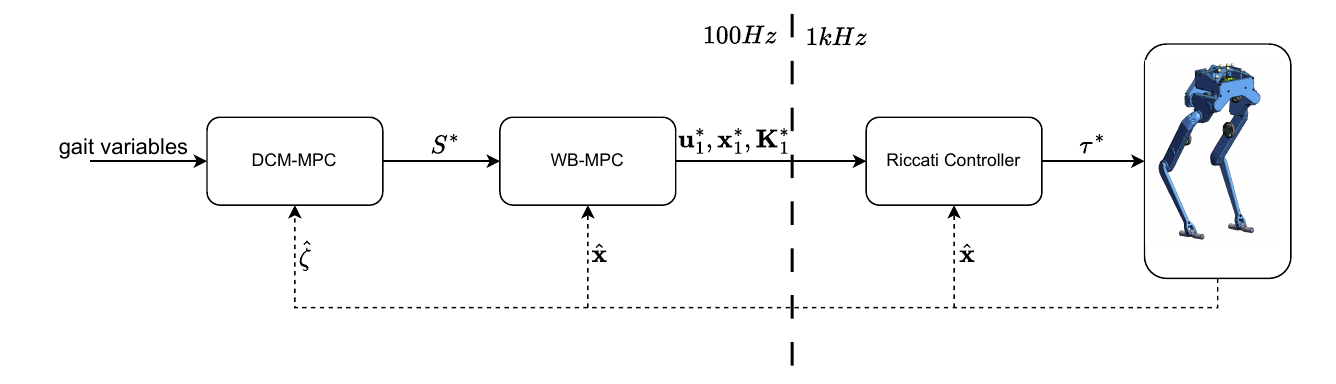}
    \caption{Simplified architecture of Bolt's walking controller.}
    \label{fig:bolt_pipeline}
\end{figure*}

\subsection{Discretized Optimal Control Formulation}
Given the step sequence generated by the sequencer, we now aim to compute a torque sequence to control the robot.
To account for all dynamic effects, we employ a whole-body MPC, which iteratively solves the Optimal Control Problem (OCP) described as follows:
\begin{equation}
    \begin{aligned}
    \min_{\mathbf{x}, \mathbf{u}} \quad & \sum_{i=0}^{N-1}l_i(\mathbf{x_i}, \mathbf{u_i})+l_N(\mathbf{x_N})\\
    \textrm{s.t.} \quad & \mathbf{x}_0=\mathbf{x}_{init} \\
    &f_i(\mathbf{x_i}, \mathbf{u_i})=\mathbf{x_{i+1}} \quad \forall i \in [0,N-1] \\
    \end{aligned}
\end{equation}
where \(\mathbf{x}\) and \(\mathbf{u}\) represent the decision variables for the state and control inputs, respectively.
\(N\) denotes the discrete horizon length (discretization of \(H_u\)), while \(l_i(\mathbf{x}, \mathbf{u})\) corresponds to the running costs and \( l_N(\mathbf{x_N}) \) the terminal cost.
The initial state, \(\mathbf{x}_0\), is set to \(\mathbf{x}_{\text{init}}\), and \(f_i(\mathbf{x}_i, \mathbf{u}_i)\) represents the system dynamics from time \(i\) to \(i+1\), including the contact dynamics constraints for both the left and right foot. The specific time instances $i$ for these contacts are provided for each foot by the step sequencer.

\subsection{Costs functions}
The running cost \( l_i(\mathbf{x_i}, \mathbf{u_i}) \) is the sum of the following costs:
\begin{enumerate}
    \item \textit{Foot height tracking cost}: The cost of foot height tracking ensures that the robot lifts its foot appropriately. It is described by the following equation:
    \begin{equation}
            \ell^f_{track}(\mathbf{x})=\left \| x_z^f - {x_z^f}^* \right \|_{w_{track}}^2
    \end{equation}
    where \( f \) denotes the flying foot, \( \mathbf{x} \) the state of the robot, \( x_z^f, ~{x_z^f}^* \in \mathbb{R} \) respectively the measured and desired flying foot height, and \( {w_{track}} \in \mathbb{R}\) a weighting hyperparameter.
    
    The reference trajectory of the foot height is a polynomial function ensuring that the height and the velocity of the foot height are zero at both the initial and final instants, and that the foot reaches a predefined height at the midpoint of the duration, as described by:
    \begin{equation}
        {x_z^f}^*(t)=\frac{16H}{t_f^4}t^4-\frac{32H}{t_f^3}t^3+\frac{16H}{t_f^2}t^2
    \end{equation}
    where \( t \in \mathbb{R} \) is a variable ranging from 0 to \( t_f \in \mathbb{R} \), \( t_f \) is the total duration of the step, and \( H \in \mathbb{R}\) is the maximum foot height at \( \frac{t_f}{2} \).
    
    Note that the foot contact positions calculated by the sequencer are not predefined; they naturally emerge from the solver used to address the WB MPC problem.
    
    \item \textit{Regularization costs}:
     A regularization cost of the control around zero command is added to indirectly limit the latter.
     It is implemented according to the following equation:
    \begin{equation}
        \ell_\mathbf{u}(\mathbf{u})=\left \| \mathbf{u} \right \|^2_{w_\mathbf{u}}
        \label{equ:reg-cost-ctrl}
    \end{equation}
    where \( {w_\mathbf{u}} \in \mathbb{R}\) is a weighting hyperparameter.
    
    Additionally, a regularization cost of the state around a reference posture is added to ensure that the robot's posture tends towards it. The cost is described as:
    \begin{equation}
        \ell_X(\mathbf{x})=\left \| (\mathbf{x}-\mathbf{x}^*) \right \|^2_{w_\mathbf{x}}
        \label{equ:reg-cost-state}
    \end{equation}
    where \( \mathbf{x}^* \in \mathbb{R}^{n_{\mathbf{x}}} \) (where \( n_{\mathbf{x}} \) is the size of the robot state vector) is the desired reference posture, and \( w_{\mathbf{x}} \in \mathbb{R}^{n_{\mathbf{x}} \times n_{\mathbf{x}}} \) is a weighting hyperparameter.
    
    \item \textit{Boundary cost}:
    A barrier cost is added to the control to ensure that it never exceeds the physical limits of the robot's actuators. The cost is given as follows:
    \begin{equation}
        \ell^f_{bu}(\mathbf{x})=\left\| \min(\max(\mathbf{u}, \bar{\mathbf{u}}), \underline{\mathbf{u}}) \right\|^2_{w_{bu}}
        \label{equ:barrier-cost-ctrl}
    \end{equation}
    where \( \underline{\mathbf{u}} \in \mathbb{R}^{n_{\mathbf{u}}} \) (where \( n_{\mathbf{u}} \) is the size of the robot control vector) and \( \bar{\mathbf{u}} \in \mathbb{R}^{n_{\mathbf{u}}} \) are respectively the lower and upper boundaries of the robot torque outputs. \( w_{bu} \in \mathbb{R} \) is a weighting hyperparameter.
\end{enumerate}

The cost \( l_N(\mathbf{x_N}) \) is the sum of the previous state-dependent costs along with an additional cost on the position of the center of mass:\begin{equation}
        \ell_{\text{CoM}}(\mathbf{x})= \left \| \mb{c} - \mb{c}^* \right \|^2_{w_\text{CoM}}
\end{equation}
where \( \mb{c} \in \mathbb{R}^3 \) is the position of the CoM, \( \mb{c}^* \in \mathbb{R}^3 \) is the reference position of the CoM, and \( w_{\text{CoM}} \in \mathbb{R} \) is a weighting hyperparameter.

The reference for the CoM is calculated by solving Eq. \eqref{equ:CoM-dynamics}:
\begin{equation}
    \mb{c}^* = (\hat{\mb{c}}_{K-1} - \hat{\zeta}_{K-1})e^{w_0(T_{K-1} - H_u)}+\hat{\zeta}_{K-1}
\end{equation}
where the subscript \(_{K-1}\) represents the index of the last step whose contact is before the horizon of the list \(S\).
This ensures that the robot's center of mass reaches a stable position while adhering to the desired setpoint.

\subsection{Control pipeline}
The full control structure consists of two parallel processes (see Fig.~\ref{fig:bolt_pipeline}):
\begin{itemize}
    \item The high-level process runs at 100 Hz and solves the DCM-MPC and WB-MPC using OSQP \cite{osqp} and Crocoddyl\cite{mastalli20crocoddyl}, respectively.
    The WB-MPC uses a 10 ms time discretization interval to balance problem-solving time and numerical integration quality.
    The high-level frequency is set to use the solution from node 1 (after 10 ms) to account for computational delay.
    
    \item The low-level process uses Riccati gains to control the robot at 1 kHz. The control law, derived from \cite{Dantec:ral:2022}, is given by:
    \begin{equation}
        \mathbf{\tau}=\mathbf{u}_1 + K_1(\hat{\mathbf{x}}-\mathbf{x_1})
    \end{equation}
    where \( \mathbf{u}_1 \in \mathbb{R}^{n_{\mathbf{u}}} \) and \( \mathbf{x}_1 \in \mathbb{R}^{n_{\mathbf{x}}} \) are the optimal control input and state computed by the solver, respectively.
    \(\hat{\mathbf{x}} \in \mathbb{R}^{n_{\mathbf{x}}} \) is the robot's measured state at 1 kHz.
    Finally, the matrix \( K_1 \in \mathbb{R}^{6 \times (n_{\mathbf{x}}-1)} \) is the Riccati gain matrix at node 1 given by:
    \begin{equation}
        K_1 \triangleq 
        \left.\begin{matrix}
        \frac{\partial \mathbf{u}}{\partial \mathbf{x}}
        \end{matrix}\right|_{\mathbf{x}_1}
\end{equation}
\end{itemize}
This setup ensures high-frequency feedback and robust real-time control.

\section{Simulation Results}

\subsection{Setup}
The objective of this simulation is to evaluate the robot's ability to follow a given velocity command, demonstrate robustness against external perturbations, and adapt its gait dynamically.
The simulation is conducted in PyBullet on an Apple Mac M3 MAX CPU clocked at 4.05 GHz with 16 cores.
A simplified URDF model of the Bolt robot is utilized\footnote{\url{https://github.com/Gepetto/example-robot-data/tree/master/robots/bolt_description}}, with details in the associated paper \cite{Grimminger2020}.
Initial conditions are identical for all simulations: Bolt starts with both feet on the ground and in the configuration specified by the SRDF.
Three scenarios are designed:
\begin{enumerate}
    \item \textit{Walking with and without disturbance}: Bolt follows a walking command while experiencing occasional external perturbations.
    \item \textit{Cluttered terrain walking}: Bolt walks on a cluttered terrain.
    \item \textit{Velocity reference transition}: Bolt transitions from a velocity command to another.
\end{enumerate}

\subsection{Nominal walking with and without perturbations}
\begin{figure}
    \centering
    \begin{subfigure}[b]{\columnwidth}
        \centering
        \includegraphics[width=\linewidth,trim={0 0.5cm 0 0},clip]{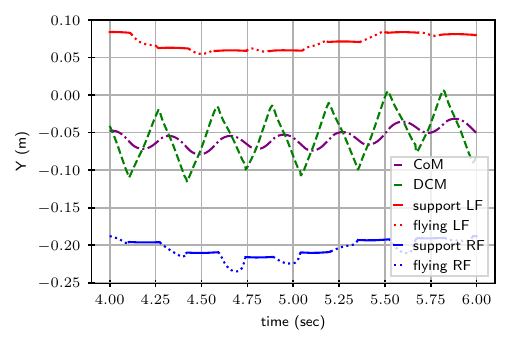}
    \end{subfigure}
    
    \begin{subfigure}[b]{\columnwidth}
        \centering
        \includegraphics[width=\linewidth]{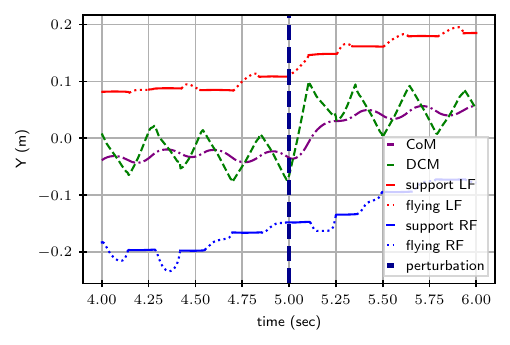}
    \end{subfigure}
    \caption{Comparison of the CoM (purple), the DCM (green), and the foot positions (red for left, blue for right) along the $y$-axis over time during walking at \( V^*_x = 0.33 \, {m.s^{-1}} \). Top: Unperturbed walking. Bottom: Walking perturbed at \( t = 5 \, \text{s} \) by an applied force of \( F = 6.3 \, \text{N} \) along the $y$-axis at the robot's base for 0.1 s.}
    \label{fig:sim_y_perturbation}
\end{figure}

We propose to compare the walking of Bolt with a target speed command \( v_X^* = 3.3 \, m.s^{-1} \) with and without perturbations. Fig.~\ref{fig:sim_y_perturbation} shows the $y$-axis positions of the COM, DCM and feet positions of bolt over time. The top plot depicts unperturbed walking over a two-second interval, while the bottom plot shows walking with a perturbation of \( 6.3 \, \text{N} \) applied at the base for \( 0.1 \, \text{s} \) (equivalent to \( 0.63 \, \text{N.s} \)) at \( t = 4 \, \text{s} \). We observe that Bolt is capable of rejecting the perturbation by adjusting the step sequence and subsequently resumes nominal walking.

\begin{figure}
    \centering
    \begin{subfigure}[b]{\columnwidth}
        \centering
        \includegraphics[width=\linewidth]{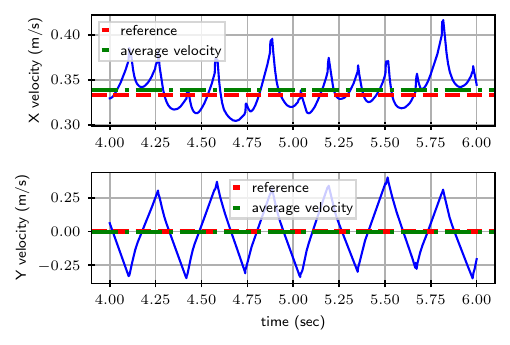}
    \end{subfigure}
    
    \begin{subfigure}[b]{\columnwidth}
        \centering
        \includegraphics[width=\linewidth]{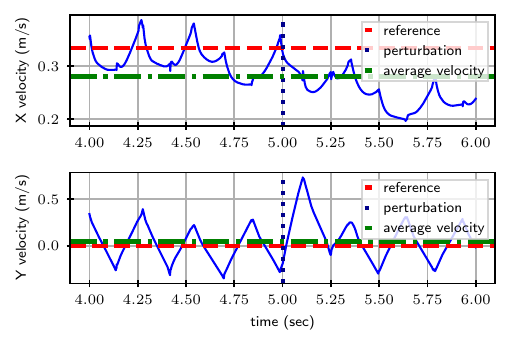}
    \end{subfigure}
    \caption{Comparison of the base velocity (blue) and the average velocity (green) along the $x$ and $y$ axes over time when walking at \( V^*_x = 0.33 \, {m.s^{-1}} \) (shown in red). Top: Unperturbed walking. Bottom: Walking perturbed at \( t = 5 \, \text{s} \) by an applied force of \( F = 6.3 \, \text{N} \) along the $y$-axis at the robot's base for 0.1 s.}
    \label{fig:sim_vel_perturbation}
\end{figure}

Next, we analyze the base velocity of Bolt.
Fig.~\ref{fig:sim_vel_perturbation} shows the base velocities in the $x$ and $y$ directions over time, both without and with the same perturbation as before.
The top plot presents the unperturbed velocities, while the bottom plot shows the velocities under perturbation.
We observe that the quality of the $x$-axis velocity tracking decreases at the moment of the perturbation and returns to the target speed once the perturbation is rejected.
This behavior is expected because, as explained in Section \ref{subsec:sensitivity_analysis}, in case of a DCM perturbation, the solver prioritizes maintaining Bolt's balance at the cost of velocity tracking.
Additionally, we note on the top plot an average error \( \mb{\bar{\epsilon_V}} = \left | \bar{\mb{V}} - \mb{V}^*  \right | = \begin{pmatrix} 5.2 & 1.1 \end{pmatrix}^\top \) \( mm.s^{-1}\) in the velocity tracking.
It is mainly due to discrepancies between the LIPM model and the simulator model, particularly the fact that angular momentum is neglected in the LIPM model.
Solutions exploring 3D DCM-based methods \cite{englsberger_three-dimensional_2015} address this issue, but it is beyond the scope of our work.

\subsection{Nominal walking on cluttered terrain}
\begin{figure}
    \centering
    \begin{subfigure}[b]{\columnwidth}
        \centering
        \includegraphics[width=\linewidth]{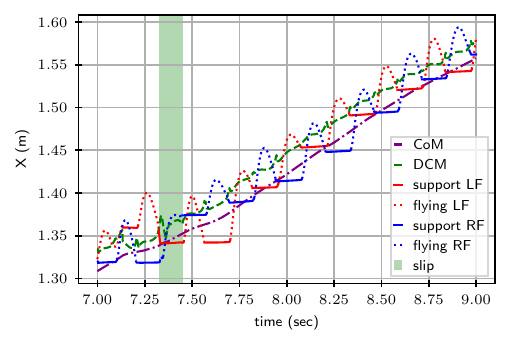}
    \end{subfigure}
    
    \begin{subfigure}[b]{\columnwidth}
        \centering
        \includegraphics[width=\linewidth]{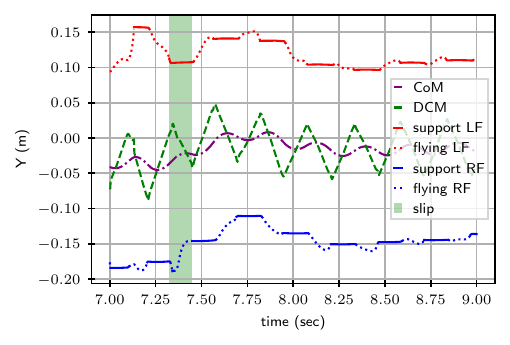}
    \end{subfigure}
    \caption{CoM (purple), the DCM (green), and the foot positions (red for left, blue for right) along the $y$-axis over time when walking at \( V^*_x = 0.33 \, {m.s^{-1}} \). The left foot is slipping in the green area.}
    \label{fig:sim_cluttered}
\end{figure}

We now demonstrate the robustness of the controller on Bolt against slips and rough terrain (see Fig.~\ref{fig:bolt_cluttered_terrain}) with a reference speed \( V_x^* = 0.3 \, m.s^{-1} \). The perturbations consist of generating rectangular parallelepipeds with side lengths ranging from 1 cm to 5 cm and heights ranging from 5 mm to 8 mm on the robot's path, with an average of 10 obstacles per square meter.
Fig.~\ref{fig:sim_cluttered} shows the $y$ axis position of the CoM, the DCM, and the foot positions over time, highlighting an instance of left foot slippage. We observe that the controller adapts the step sequence to compensate for slips and reject the resulting perturbations.

\subsection{Transition from standing to walking}
\begin{figure}
    \centering
        \includegraphics[width=\linewidth]{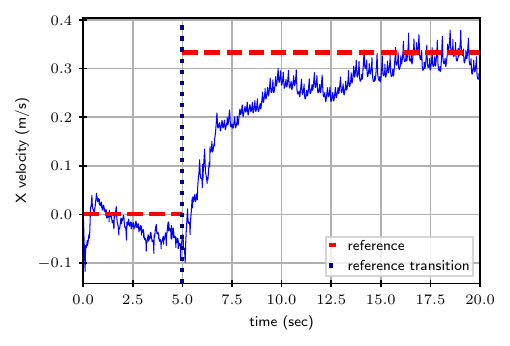}
    \caption{Base velocity (blue) and the average velocity (green) along $x$-axis over time when walking at \( V^*_x = 0.0 \, {m.s^{-1}} \) before $t=5s$ and \( V^*_x = 0.33 \, {m.s^{-1}} \) after $t=5s$ (shown in red).}
    \label{fig:vel_standing_to_walking}
\end{figure}

Fig.~\ref{fig:vel_standing_to_walking} shows the robot's base velocity along the $x$ axis over time. 
The command initially sets the velocity to $0\; m.s^{-1}$ (step in place), transitioning to $0.3\; m.s^{-1}$ at \( t = 5 \)s. 
We measure a rise time of 4.35s for the robot to reach its final velocity value. 
This delay can be attributed to a significant increase in angular moments caused by the velocity change, leading to a larger discrepancy between the LIPM and the simulator model. 
This difference can be modeled as a perturbation on the DCM. 
As explained in section \ref{subsec:sensitivity_analysis}, foot placements are more sensitive to perturbations than robot balance. 
Therefore, while striving to maintain balance, the foot positions deviate from the optimal position, resulting in a slower achievement of the desired velocity.

\section{Discussion and conclusion}
In this study, we demonstrated that successful control of a highly unstable biped robot can be achieved using a footstep sequencer and a whole-body model predictive controller (WB MPC) operating at $100~Hz$.
The specificity of our approach is that it does not require explicit footstep trajectories as they emerge naturally when solving the WB MPC.
Our results show that the robot exhibits significant capabilities in executing velocity transitions, recovering from perturbations, and managing foot slippage.
Sensitivity analysis provides additional insights, aligning with previous research while offering a deeper theoretical understanding.

However, we observed that the desired speed was not always attained and that velocity transitions were slow, which can limit the robot's performance in scenarios requiring rapid speed changes.
This outcome was anticipated due to the assumptions imposed by the LIPM.

In the short term, we plan to deploy and validate this controller on the physical Bolt robot.
For future work, we plan to leverage the sensitivity analysis computed in this study as a foundational work to estimate the sensitivity of the whole control structure (DCM MPC + WB MPC).
We would additionally need to compute the sensitivity WB MPC with respect to the contact timings calculated by the step sequencer.
This study could allow to provide more depth in the theoretical interpretation of the performance of our control structure as well as offering robustness guarantees.

\section*{Acknowledgment}
This work was supported by the cooperation agreement ROBOTEX 2.0 (Grants ROBOTEX ANR-10-EQPX-44-01 and TIRREX-ANR-21-ESRE-0015).

\printbibliography

\end{document}